%% file: main.tex
\documentclass[sw,crcready]{iosart2x}

\usepackage{dcolumn}

\newcolumntype{d}[1]{D{.}{.}{#1}}
\firstpage{1}
\lastpage{13}
\volume{1}
\pubyear{2024}
\usepackage{xcolor}
\usepackage{float}
\usepackage{booktabs}
\usepackage[capitalise,noabbrev]{cleveref}
\usepackage{graphicx}
\usepackage{grffile}
\usepackage{hyperref}
\begin{document}
\begin{frontmatter} 

%
\title{Retrieval-Augmented Generation-based Relation Extraction}

\runtitle{RAG-based Relation Extraction}


\begin{aug}
\author[A,B]{\inits{S.}\fnms{Sefika} \snm{Efeoglu}\ead[label=e1]{sefika.efeoglu@tu-berlin.de}%
\ead[label=e3]{sefika.efeoglu@fu-berlin.de}%
\thanks{Corresponding author. \printead{e1}.}}

\author[B,C]{\inits{A.}\fnms{Adrian} \snm{Paschke}\ead[label=e2]{adrian.paschke@fokus.fraunhofer.de}}

\address[A]{Electrical Engineering and Computer Science Department, Technische Universitaet Berlin, Berlin, \cny{Germany}\printead[presep={\\}]{e1}}
\address[B]{Department of Computer Science, \orgname{Freie Universitaet Berlin},
Berlin, \cny{Germany}\printead[presep={\\}]{e2}}
\address[C]{Data Analytic Center (DANA), \orgname{Fraunhofer Institute FOKUS},
Berlin, \cny{Germany}}
\end{aug}



\begin{abstract}

Information Extraction (IE) is a transformative process that converts unstructured text data into a structured format by employing entity and relation extraction (RE) methodologies. The identification of the relation between a pair of entities plays a crucial role within this framework. Despite the existence of various techniques for relation extraction, their efficacy heavily relies on access to labeled data and substantial computational resources. In addressing these challenges, Large Language Models (LLMs) emerge as promising solutions; however, they might return hallucinating responses due to their own training data. To overcome these limitations, Retrieved-Augmented Generation-based Relation Extraction (RAG4RE) in this work is proposed, offering a pathway to enhance the performance of relation extraction tasks.

This work evaluated the effectiveness of our RAG4RE approach utilizing different LLMs. Through the utilization of established benchmarks, such as TACRED, TACREV,  Re-TACRED, and SemEval RE datasets, our aim is to comprehensively evaluate the efficacy of our RAG4RE approach. In particularly, we leverage prominent LLMs including Flan T5, Llama2, and Mistral in our investigation. The results of our study demonstrate that our RAG4RE approach surpasses performance of traditional RE approaches based solely on LLMs, particularly evident in the TACRED dataset and its variations. Furthermore, our approach exhibits remarkable performance compared to previous RE methodologies across both TACRED and TACREV datasets, underscoring its efficacy and potential for advancing RE tasks in natural language processing.
\end{abstract}

\end{frontmatter}

\input{sections/introduction}
\input{sections/relatedworks}
\input{sections/methodology}
\input{sections/evaluation}

\input{sections/conclusion}

\bibliographystyle{ios1}  
\bibliography{reference}

\end{document}

%% file: sections/introduction.tex
\section{Introduction}
\begin{figure}[H]
    \centering
    \includegraphics[bb=0 0 300 40]{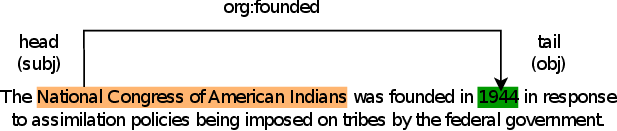}
    \caption{An example of relation types between head and tail entities in a sentence.
    }
    \label{fig:example_re}
\end{figure}

Information Extraction (IE) is a process of converting unstructured text data into structured data by applying entity and relation extraction approaches. Identifying the relation between a pair of entities in a sentence, Relation Extraction (RE), is one of the most significant tasks in the IE pipeline~\cite{Grishman_2015}. The RE has a pivotal role in constructing domain specific Knowledge Graphs (KGs) from text data, and the completeness of KGs.
An example of relation types between entity pairs, \textit{org:founded}, between head (\textit{National Congress of American Indians}) and tail (\textit{1994}) entities is represented in Figure~\ref{fig:example_re}. Various RE approaches have been developed, including supervised RE, distant supervision, and unsupervised RE methods. However, well-performing RE approaches require a large amount of labeled data and substantial computation time. Another effective method for identifying the relation types between entities is fine-tuning language models~\cite{wang-etal-2022-deepstruct,CHEN2024123478,cohen2020relation,zhou-chen-2022-improved,li2022reviewing}. However, it is worth noting that this approach requires substantial GPU memory and computational time.

Large Language Models (LLMs) possess incredible inference capabilities that enable them to address key tasks in IE, such as Entity Recognition (ER) and RE. Nevertheless, their abilities are restricted by their training data; therefore, they are prone to producing hallucinating results when lacking prior knowledge. Additionally, LLM prompt tuning approaches require prompt template engineering and domain experts in domain-specific information extraction approaches~\cite{CHEN2024123478}; however, the template engineering leads to an expensive time cost due to its manual process.
Retrieved-Augmented Generation (RAG) is proposed to reduce the hallucination of the LLMs when LLM-based conversational systems produce random responses to a query~\cite{rag_meta}. The RAG system functions akin to an open-book exam, integrating relevant information from the Embedding DB directly into the query (sentence)~\cite{rag_meta}.

Well-performing RE approaches require a large amount of labeled training data and computation time, since they learn RE patterns in a supervised manner. Fine-tuning LLMs demands a substantial amount of GPU memory and computation time when training data is large. In the era of the LLMs, well-designed prompts might help us identify the relations between entities in a sentence. It is clear that well-designed prompts yield highly accurate performance in other downstream tasks, such as ontology-driven knowledge graph generation~\cite{text_to_kg}, text-to-image generation~\cite{draw_me} and ontology matching~\cite{hertling2023olala}. Drawing inspiration from these works, enriching the context of the prompt might provide task-relevant knowledge to the LLMs and enhance their responses to a prompt.

In this work, our aim is to explore the potential enhancement in performance of relation extraction between entity pairs in the sentence through the utilization of an RAG-based RE (RAG4RE) approach~\footnote{The source code is available at~\url{https://github.com/sefeoglu/RAG4RE}}. To evaluate our RAG4RE approach, we leverage RE benchmark datasets, such as TACRED~\cite{zhang-etal-2017-position}, TACREV~\cite{alt-etal-2020-tacred}, Re-TACRED~\cite{Stoica_Platanios_Poczos_2021} and SemEval~\cite{hendrickx-etal-2010-semeval} RE datasets. We leverage encoder and decoder models like Flan T5~\cite{Flan_T5}, and decoder-only models such as Llama2~\cite{Touvron2023Llama2O} and Mistral~\cite{jiang2023mistral}, which have been integrated into our approach described in Figure~\ref{fig:re-diagram}. 
In this work, our findings are:
\begin{enumerate}
    \item The RAG-based RE approach has the potential to outperform both simple query (without relevant sentence), known as Vanilla LLM prompting, and existing best-performing RE approaches from previous studies.
    \item While Decoder-only LLMs~\cite{pan2024unifying} still encounters hallucination issues on these datasets, our RAG-based RE approach effectively mitigates this problem, especially when compared to the results obtained from the simple query.
\end{enumerate}

In the following section, we first summarize recent works in RE and RAG in the Related Works section, and then introduce a detailed description of the proposed RAG4RE approach in the Methodology section. We evaluate our RAG4RE approach on RE benchmark datasets along with integrating different types of LLMs and discuss their results in comparison with those of previous approaches in the Evaluation section. Finally, we summarize the outcomes of our RAG4RE approach in the Conclusion section.

%% file: sections/relatedworks.tex
\section{Related Works}
\label{sec:relatedwork}
We summarize recent works into two categories: (i) Relation Extraction and (ii) Retrieval-Augmented Generation.
\subsection{Relation Extraction}
Relation Extraction (RE) is one of the main tasks of Information Extraction and plays a significant role among natural language processing tasks. RE aims to identify or classify the relations between entity pairs (head and tail entities).

RE can be carried out with various types of approaches: (i) supervised techniques including features-based and kernel-based methods, (ii) a special class of techniques which jointly extract entities and relations (semi-supervised), (iii) unsupervised, (iv) Open IE and (v) distant supervision-based techniques~\cite{Sachin_2017}. Supervised techniques require a large annotated data set, and its annotation process is time-consuming and expensive~\cite{Sachin_2017}. Distant supervision is amongst one of the popular methods dealing with this annotated data problem. The distant supervision, based on existing knowledge bases, brings its own drawback, and it faces the issue of wrongly labeled sentences troubling the training due to the excessive amount of noise~\cite{Aydar_2020}. Another popular approach is weakly supervised RE~\cite{Eugene_2000}. However, the weakly supervised approach is more error-prone because of semantic drift in a set of patterns per iteration of its incremental learning approach like a snowball algorithm~\cite{Eugene_2000}. In rule-based RE approaches, finding relations is mostly restricted by predefined rules~\cite{Sachin_2017}.

In terms of the best-performing RE approach, obtained by fine-tuning the language models, Cohen et al.~\cite{cohen2020relation} proposed a span-prediction-based approach for relation classification instead of single embedding to represent the relation between entities. This approach has been improved the state-of-the-art scores on the well-known datasets. DeepStruct~\cite{wang-etal-2022-deepstruct} proposed an innovative approach aimed at enhancing the structural understanding capabilities of language models. This work introduced a pretrained model comprising 10 billion parameters, facilitating the seamless transfer of language models to structure prediction tasks. Specifically, regarding the RE task, the output format entails a structured representation of (head entity, relation, tail entity), while the input format comprises the input text along with a pair of head and tail entities. Zhou et al.~\cite{zhou-chen-2022-improved} concentrated on addressing two critical issues that affect the performance of existing sentence-level RE models: (i) Entity Representation and (ii) noisy or ill-defined labels. Their approach extends the pretraining objective of masked language modeling to entities and incorporates a sophisticated entity-aware self-attention mechanism, enabling more accurate and robust RE. Li et al.~\cite{li2022reviewing} proposed a label graph to review candidate labels in the top-K prediction set and learn the connections between them. When predicting the correct label, they first compute that the top-K prediction set of a given sample contains useful information.

Zhang et al.~\cite{Zhang2023LLM-QA4RE} generated multiple-choice question prompts from test sentences where choices consist of verbalizations of entities and possible relation types. These choices are selected from the training sentence based on entities in a test sentence. However, it could not outperformed the previously introduced rule and ML- based approaches. In the context of their works, Zhang et al.~\cite{Zhang2023LLM-QA4RE} proved that enriching prompt context improves the prediction results on benchmark datasets such as TACRED and Re-TACRED. Melz~\cite{melz2023enhancing} focuses on Auxiliary Rationale Memory for the RAG approach in the Relation Extraction task, and the proposed system learns from its successes without incurring high training costs. Chen et al.~\cite{CHEN2024123478} proposes a Generative Context-Aware Prompt-tuning method which also tackles the problem of prompt template engineering. This work proposed a prompt generator that is used to find context-aware prompt tokens by extracting and generating words regarding entity pairs and evaluated on four benchmark datasets: TACRED, TACREV, Re-TACRED and SemEval.

\subsection{Retrieval-Augmented Generation}
Retrieval-Augmented Generation (RAG) for large language models can be classified into two categories: i) naive RAG and ii) advanced RAG. Naive RAG has basic steps: retrieve, augmentation, and generation, while the advanced version includes a post-processing step before sending the retrieved information to a user~\cite{gao2023retrieval}.
The concept of RAG has been suggested as a way to minimize the undesired alterations in Language Models (LLMs) when conversational systems built on LLMs generate arbitrary responses to a query~\cite{rag_meta}. RAG is an example of open-book exams which are applied to the usage of LLMs. The retriever mechanism in RAG finds an example of the user query (prompt), and then the user query is regenerated along with the example by the data-augmentation module in RAG. Ovadia et al.~\cite{ovadia2023fine} evaluates the knowledge injection capacities of both fine-tuning and the RAG approach and found that LLMs dealt with performance problems through unsupervised fine-tuning while RAG outperformed the fine-tuning approach in unsupervised learning.

In this work, we introduce a Retrieval-Augmented Generation-based Relation Extraction (RAG4RE) approach to identify the relationship between a pair of entities in a sentence.

%% file: sections/methodology.tex
\section{Methodology}
\begin{figure}[H]
    \centering
    \includegraphics[bb=0 0 1800 300, height=0.5\linewidth]{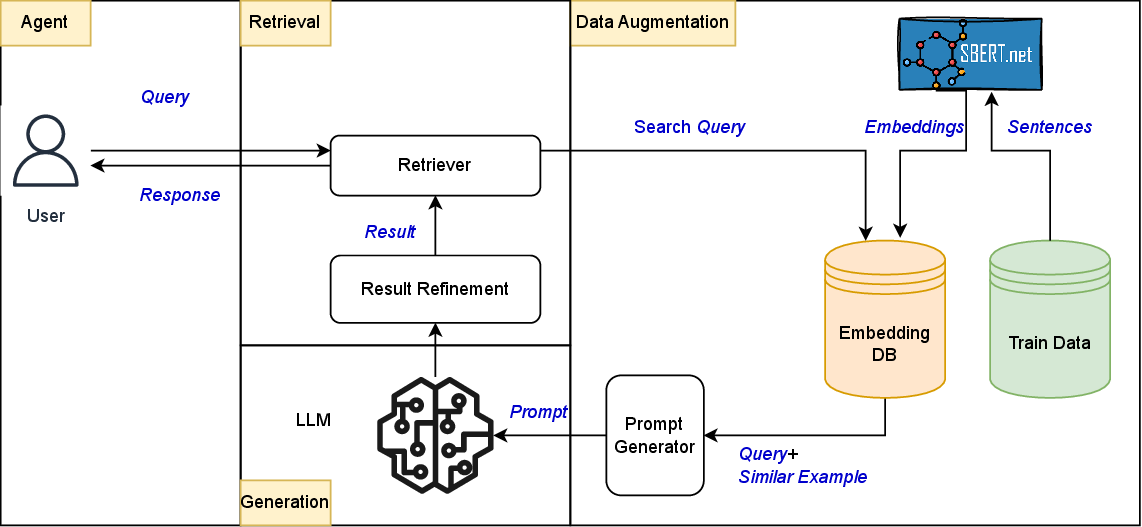}
    \caption{RAG-based Relation Extraction pipeline.}
    \label{fig:re-diagram}
\end{figure}

In this work, we have developed an RAG-based Relation Extraction (RAG4RE) approach to identify the relation between a pair of entities in a sentence. Our proposed RAG4RE, illustrated in Figure~\ref{fig:re-diagram}, consists of three modules: (i) Retrieval, (ii) Data Augmentation, and (iii) Generation. Our proposed RAG4RE approach is a variant of an advanced RAG~\cite{gao2023retrieval}, as its retrieval module includes ``Result Refinement'' which applies post-processing after responses from generation module. An example of how our RAG4RE returns different responses to both RAGRE and a simple query can be seen in~\Cref{tab:retrieval_exp}

The rest of the section explains the details of how each module of our proposed approach at~\Cref{fig:re-diagram} works under specific subsections. 

\subsection{Retrieval}

A user submits a sentence (query) along with a pair of entities (head and tail entities) that might have a relation to the Retrieval module. Then, the Retriever sends this query to the Data Augmentation module, which extends the original query with a semantically similar sentence from training dataset, as an example given in~\Cref{fig:prompt_example}. ``Result Refinement'' in this module applies post-processing approaches when results are returned from the Generation module. The ``Result Refinement'' consists of a couple of response processing steps, such as refining prefixes (e.g., changing ``founded'' to ``org:founded''), and converting ``no relation'' answers into ``no\_relation'' as defined in the predefined relation types. This ensures that only relation types are returned instead of explanations from the LLMs.

\subsection{Data Augmentation}

\begin{figure}[H]
    \centering
    \includegraphics[bb=10 0 200 120, height=0.2\linewidth]{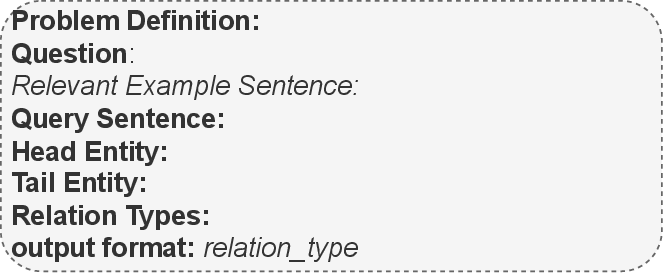}
    \caption{Re-generated prompt template}
    \label{fig:prompt_tempt}
\end{figure}

\begin{figure}[H]
    \centering
    \includegraphics[bb=0 0 400 300, height=0.5\linewidth]{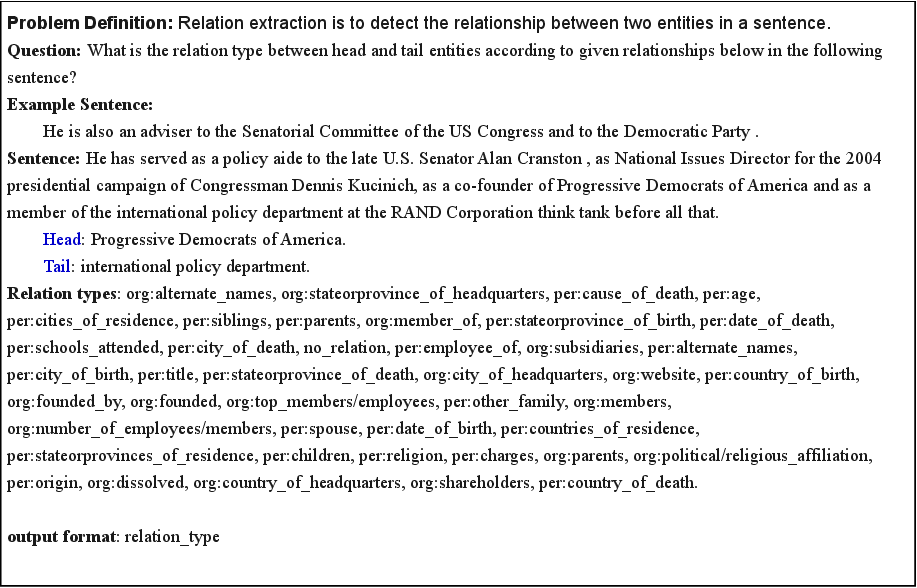}
    \caption{A regenerated prompt is illustrated.}
    \label{fig:prompt_example}
\end{figure}

The Data Augmentation module contains an Embedding Database (DB) consisting of embeddings of training data, computed using the Sentence BERT (SBERT) model~\cite{reimers2019sentence}. We used  the ``all-MiniLM-L6-v2''~\footnote{The model is available at~\url{https://huggingface.co/sentence-transformers/all-MiniLM-L12-v2}} version of SBERT in our approach. Additionally, the query embedding is computed by SBERT within this module. Subsequently, the system calculates similarity scores between the embeddings of each line of training sentences in the Embedding DB and the query embeddings, utilizing the cosine similarity metric. The system considers the most similar embeddings from the Embedding DB as an example of the query. Both the query and the similar sentence example are then inputted into the prompt generator, which prepares prompts by considering the template defined in~\Cref{fig:prompt_tempt}. In other words, prompt generator regenerates the user query along with the relevant sentence that was founded at the Embedding DB. One of the generated prompts can be seen in~\Cref{fig:prompt_example}. The prompt generator then inputs the prompt (regenerated user query), consisting of the query and its similar example, into the Generation module.

\subsection{Generation}

The LLM generates a response for the prompts using a zero-shot prompting approach in the Generation module. We integrate LLMs with different architectures that are encoder-and-decoder and decoder-only~\cite{pan2024unifying} in our experiments so that we can evaluate the performance of our proposed RAG4RE approach with  the different LLMs and compare themselves within the RAG4RE approach. Subsequently, the response is sent to the ``Result Refinement" in the Retrieval module. Result refinement becomes necessary when the response might include a missing prefix, such as ``founded'' instead of ``org:founded''. The Generation module concludes when the results are sent into the Retrieval. Afterwards, the ``Retriever'' sends the results to the user.

%% file: sections/evaluation.tex
\section{Evaluation}
In this section, we examine performance of our RAG4RE work. We first introduce experimental settings in~\Cref{sec:experimental_setup} and then give results of our experiments in~\Cref{sec:Results}. Afterwards, false predictions are analyzed in~\Cref{sec:err_analysis}. Lastly, we discuss the performance of our approach with previous works.

\subsection{Experimental Setup}
\label{sec:experimental_setup}
In our work, we assess the effectiveness of our RAG4RE approach by leveraging well-established RE benchmarks, including TACRED~\cite{zhang-etal-2017-position}, TACREV~\cite{alt-etal-2020-tacred}, Re-TACRED~\cite{Stoica_Platanios_Poczos_2021}, and SemEval~\cite{hendrickx-etal-2010-semeval} RE datasets. These benchmarks include head and tail entities in given sentences, as well as ground truth relation types between these entities. These widely recognized datasets serve as invaluable resources for assessing the efficiency and performance of our approach. Further insights into the dataset can be seen below, and in~\Cref{tab:dataset}.
\begin{itemize}
    \item \textbf{TACRED}, namely the TAC RE Dataset, is a supervised RE dataset obtained via crowdsourcing and targeted towards TAC KBP relations. There are no direction of predefined relations, which can also be extracted from a given sentence tokens. We directly used this licensed dataset from Linguistic Data Consortium (LDC)~\footnote{The TAC Relation Extraction dataset catalog is accessible at~\url{https://catalog.ldc.upenn.edu/LDC2018T24}}.\\
    \item \textbf{TACREV} is a revisited version of TACRED that reduces noise in sentences defined with ``no\_relation''. In our work, this dataset is generated from original TACRED by running source codes given at~\cite{alt-etal-2020-tacred}~\footnote{The TACREV source code repository is available at~\url{https://github.com/DFKI-NLP/tacrev}.}.\\
    \item \textbf{Re-TACRED} is a re-annotated version of the TACRED dataset that can be used to perform reliable evaluations of RE models. To generate this Re-TACRED from original TACRED, we leverage source codes~\footnote{The Re-TACRED source code are at~\url{https://github.com/gstoica27/Re-TACRED}.} given at~\cite{Stoica_Platanios_Poczos_2021}.\\
    \item \textbf{SemEval:} focuses on multi-way classification of semantic relationships between entity pairs. The predefined relations (target relation labels) in this benchmark dataset have directions and cannot be extracted from tokens of (test, or train) sentences in the dataset. The dataset is used from HuggingFace~\footnote{The SemEval Dataset card:~\url{https://huggingface.co/datasets/sem_eval_2010_task_8}.}.
\end{itemize}
\begin{table}[H]
    \centering
        \caption{Overview of benchmark datasets.}
    \label{tab:dataset}
    \begin{tabular}{c|c|c|c|c}
    \toprule
    Split&\textbf{TACRED}&\textbf{TACREV}&\textbf{Re-TACRED}&\textbf{SemEval}\\
    \midrule
    Training&68124&68124&58465&8000\\
    Test&15509&15509&13418&2717\\
    Validation&22631&22631&19584&-\\
    \textit{\#} of Relations &42&42&40&19\\
    \bottomrule
    \end{tabular}
    \end{table}

Additionally, we evaluate our proposed method on these datasets by integrating various LLMs, such as Flan T5 (XL and XXL), Llama2-7b, and Mistral-7B-Instruct-v0.2 models. The prompt template for our RAG4RE includes a relevant sentence provided by the user (See the prompt template in~\Cref{fig:prompt_tempt}), whereas the simple query does not include any relevant sentence regarding the query. This comprehensive evaluation helps us understand the performance of our approach across various LLMs. Lastly, we compare our RAG4RE approach with simple query (Vanilla LLM prompting) in terms of micro F1, Recall, and Precision scores, since these benchmark datasets are imbalance.
To compute these metrics, we leverage metrics library of sklearn~\footnote{Sklearn evaluation library: ~\url{https://scikit-learn.org/stable/modules/model_evaluation.html}}.

In regard to hardware specifications, these language models have undergone evaluation utilizing a setup comprising 4 GPUs, with each GPU boasting a memory capacity of 12 GB in the NVIDIA system. Furthermore, the memory configuration reaches 300 GB.

\input{sections/results}

\input{sections/discussion}

%% file: sections/results.tex
\subsection{Results}
\label{sec:Results}
Our experiments were conducted using the four benchmark datasets mentioned above. Firstly, we assessed the performance of our proposed RAG4RE and compared it to that of a simple query (sentence), Vanilla LLM prompting, in terms of micro F1 score. As mentioned earlier, our evaluation criteria took into account the micro F1 score, Recall, and Precision due to the imbalanced labelling of the datasets.
Furthermore, we explored how our approach enhances the performance of LLM responses. This was achieved by incorporating a similar example sentence into the prompt template alongside the query sentence in our proposed RAG4RE approach (See Data Augmentation at~\Cref{fig:re-diagram}). We compared the outcomes of a simple query without any relevant sentence to our results (See~\Cref{tab:sota_results} and~\Cref{tab:error_analysis}).

\begin{figure}[H] 
    \caption{Micro F1 scores of four different benchmark datasets.}
    \label{fig:ex_results} 
  \centering
  \begin{minipage}[b]{0.61\linewidth}
    \includegraphics[bb=0 0 200 300, height=0.5\linewidth]{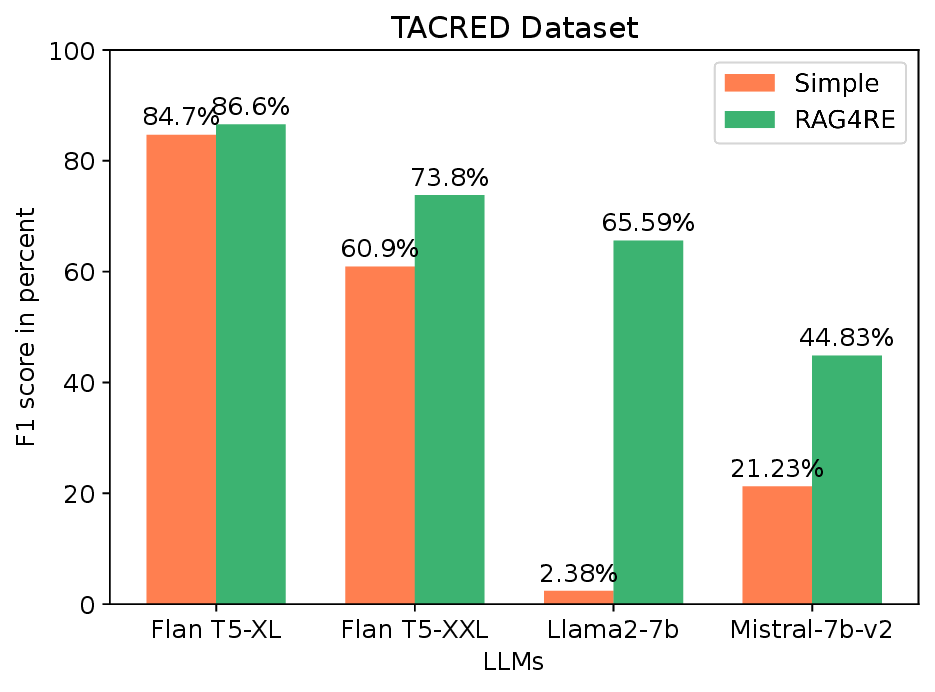} 
    \vspace{4ex}
  \end{minipage}
  \begin{minipage}[b]{0.61\linewidth}
    \includegraphics[bb=0 0 200 300, height=0.5\linewidth]{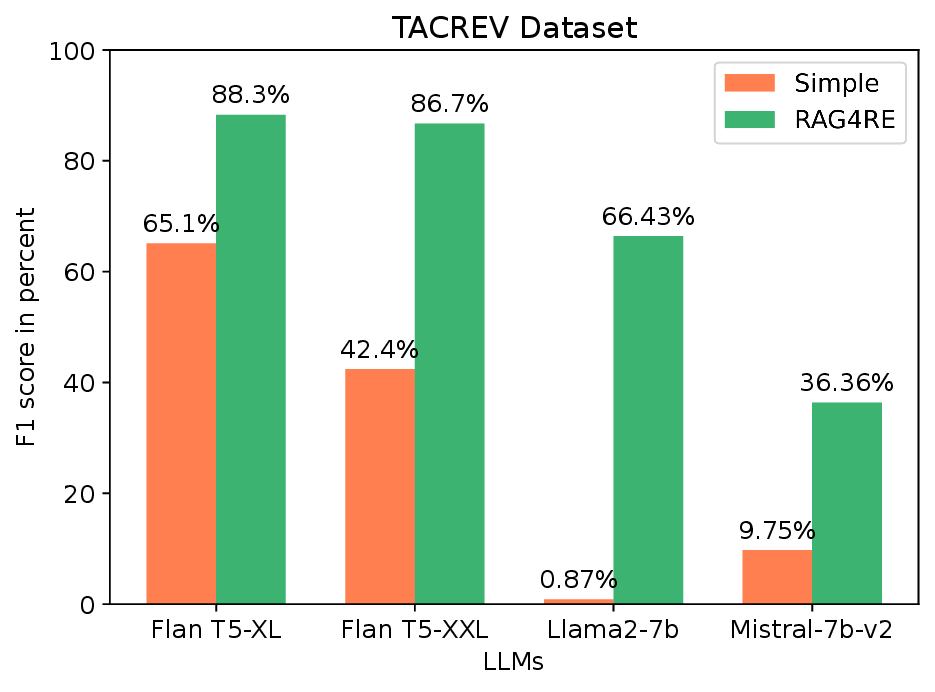} 
    \vspace{4ex}
  \end{minipage} 
  \begin{minipage}[b]{0.61\linewidth}
    \includegraphics[bb=0 0 200 300, height=0.5\linewidth]{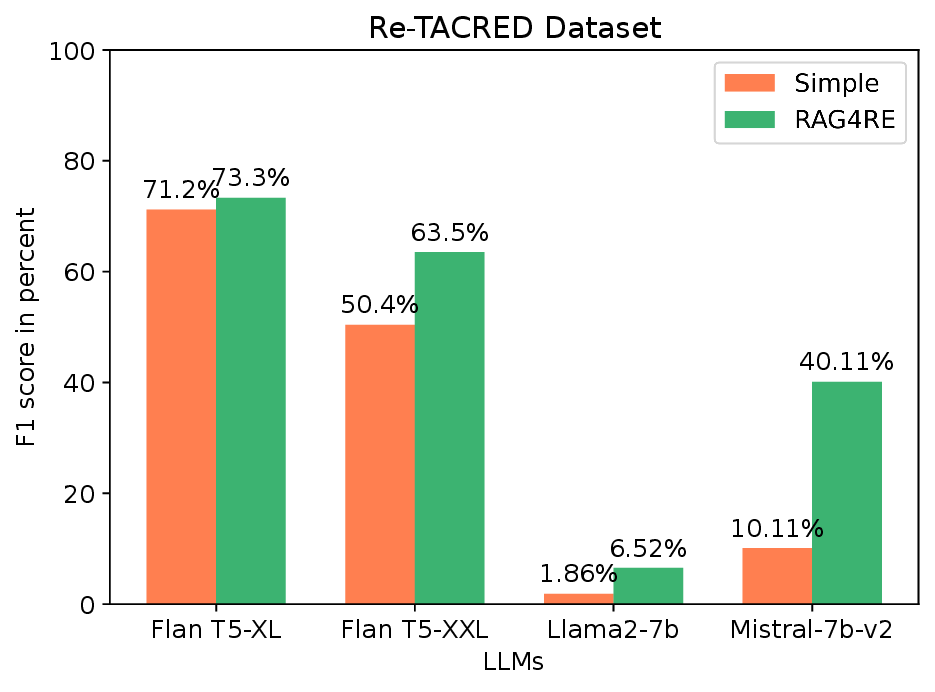} 
    \vspace{4ex}
  \end{minipage}
  \begin{minipage}[b]{0.61\linewidth}
    \includegraphics[bb=0 0 200 300, height=0.5\linewidth]{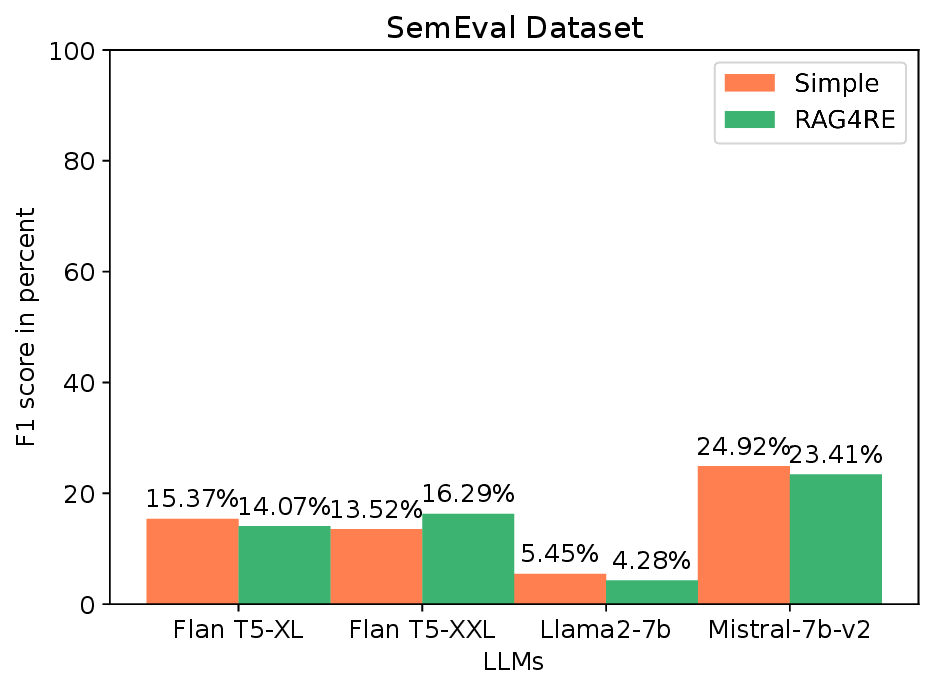} 
    \vspace{4ex}
  \end{minipage}
\end{figure}

We leveraged various LLMs to conduct our experiments and evaluated the performance of our proposed RAG4RE. Our RAG4RE has achieved remarkable results when compared its performance to that of the simple query at~\Cref{tab:sota_results} and~\Cref{fig:ex_results}. It can be seen that our RAG4RE outperforms the simple query sending to the LLMs in TACRED, TACREV, and Re-TACRED despite the fact that the language model has been changed. The highest F1 scores amongst the results at~\cref{tab:sota_results} have been accomplished 86.6\%, 88.3\% and 73.3\% of F1 scores on the TACRED, TACREV and Re-TACRED, respectively. These outstanding scores were achieved through the integration of the Flan T5-XL model into the Generation module. 
Unfortunately, our RAG4RE fell short of achieving comparable performance on the SemEval dataset. This is primarily because the predefined relations (target relation labels) within this dataset cannot be directly extracted from the sentence tokens themselves.

The notable improvement observed in RAG4RE's results can be primarily attributed to the incorporation of a relevant example sentences extracted from benchmark datasets' training sentences into the user query. As pointed out in~\cite{rag_meta}, RAG operates akin to an open book exam, and adding the relevant sentence regarding a query sentence plays a facilitating role in understanding of the query by the LLMs in the Generator module of our approach at~\Cref{fig:re-diagram}. This understanding is supported by the results outlined in~\Cref{tab:sota_results}.

\begin{table}[H]
  \centering
  \caption{Results of the experiments conducted on four different benchmark datasets alongside different LLMs.}
  \label{tab:sota_results}
  \begin{tabular}{p{1.5cm} |l| r r r|r r r|r r r |r r r}
    \toprule
     \multicolumn{2}{c}{\textbf{}} & \multicolumn{3}{c}{\textbf{TACRED}} &\multicolumn{3}{c}{\textbf{TACREV}}&\multicolumn{3}{c}{\textbf{Re-TACRED}} &\multicolumn{3}{c}{\textbf{SemEval}} \\
    \cmidrule{1-14}
    \multicolumn{1}{c}{\textbf{LLM}} & \multicolumn{1}{c}{\textbf{Approach}} & \multicolumn{1}{c}{P(\%)} & \multicolumn{1}{c}{R(\%)}& \multicolumn{1}{c}{F1(\%)} &\multicolumn{1}{c}{P(\%)} & \multicolumn{1}{c}{R(\%)}& \multicolumn{1}{c}{F1(\%)}&\multicolumn{1}{c}{P(\%)} & \multicolumn{1}{c}{R(\%)}& \multicolumn{1}{c}{F1(\%)} &\multicolumn{1}{c}{P(\%)} & \multicolumn{1}{c}{R(\%)}& \multicolumn{1}{c}{F1(\%)}\\
    \midrule
    \textit{Flan T5-XL} &simple query  &$91.0$ & $79.2$ & $84.7$ &$97.2$ &$49.0$ & $65.1$&$69.5$ &$73.0$ &$71.2$&$20.41$ & $12.32$ & $15.37$\\
      & RAG4RE &$84.5$&$88.8$&\textbf{86.6}&$84.5$ &$92.4$&$\textbf{88.3}$&$63.0$&$87.7$&$\textbf{73.3}$ &$17.16$ & $11.93$ &$14.07$\\
    \hline
   \textit{Flan T5-XXL}& simple query &$94.9$&$44.7$&$60.9$&$98.2$&$27.1$&$42.4$&$70.2$&$39.3$&$50.4$& 13.64 &13.4 & 13.52\\
      & RAG4RE &$93.3$&$61.0$&$73.8$&$91.7$ &$82.3$&$86.7$&$78.0$&$53.6$&$63.5$&17.32 & 15.39 & 16.29\\
      \hline
      
      \textit{Llama2-7b}&simple query &84.97&1.21&2.38&74.64&0.44&0.87 &80.2&0.94&1.86& 5.89 & 5.08 & 5.45 \\
      & RAG4RE &81.23  &55.01 &65.59 &84.89 &54.57&66.43 &55.93& 3.46& 6.52& 4.36 & 4.2 &4.28\\
      \hline
    \textit{Mistral-7B-Instruct-v0.2} & simple query& 94.67&11.96&21.23&92.34&5.15&9.75&64.64&5.48&10.11 &25.5 & 24.37 & 24.92 \\
    
         & RAG4RE &87.81&30.1&44.83&93.23&22.59&36.36&60.19&30.08&40.11 & 24.1 & 22.75 & 23.41\\
    \bottomrule
  \end{tabular}
\end{table}

Consequently, it is clear that our RAG4RE approach, consisting of the relevant example sentence about the query, has improved F1 scores on  benchmark datasets whose predefined relations (target relationship labels) are coming from the given sentence tokens e.g., TACRED, TACREV and Re-TACRED, when compared its results to those of  a simple query (Vanilla LLM prompting) as demonstrated in~\Cref{tab:sota_results} and~\Cref{fig:ex_results}.



\subsection{Error Analysis}
\label{sec:err_analysis}
\begin{figure}[H] 
    \caption{The number of False Negatives and Positives in results of experiments conducted on different benchmark datasets.}
    \label{fig:err} 
  \centering
  \begin{minipage}[b]{0.62\linewidth}
    \includegraphics[bb=0 0 200 500, height=0.5\linewidth]{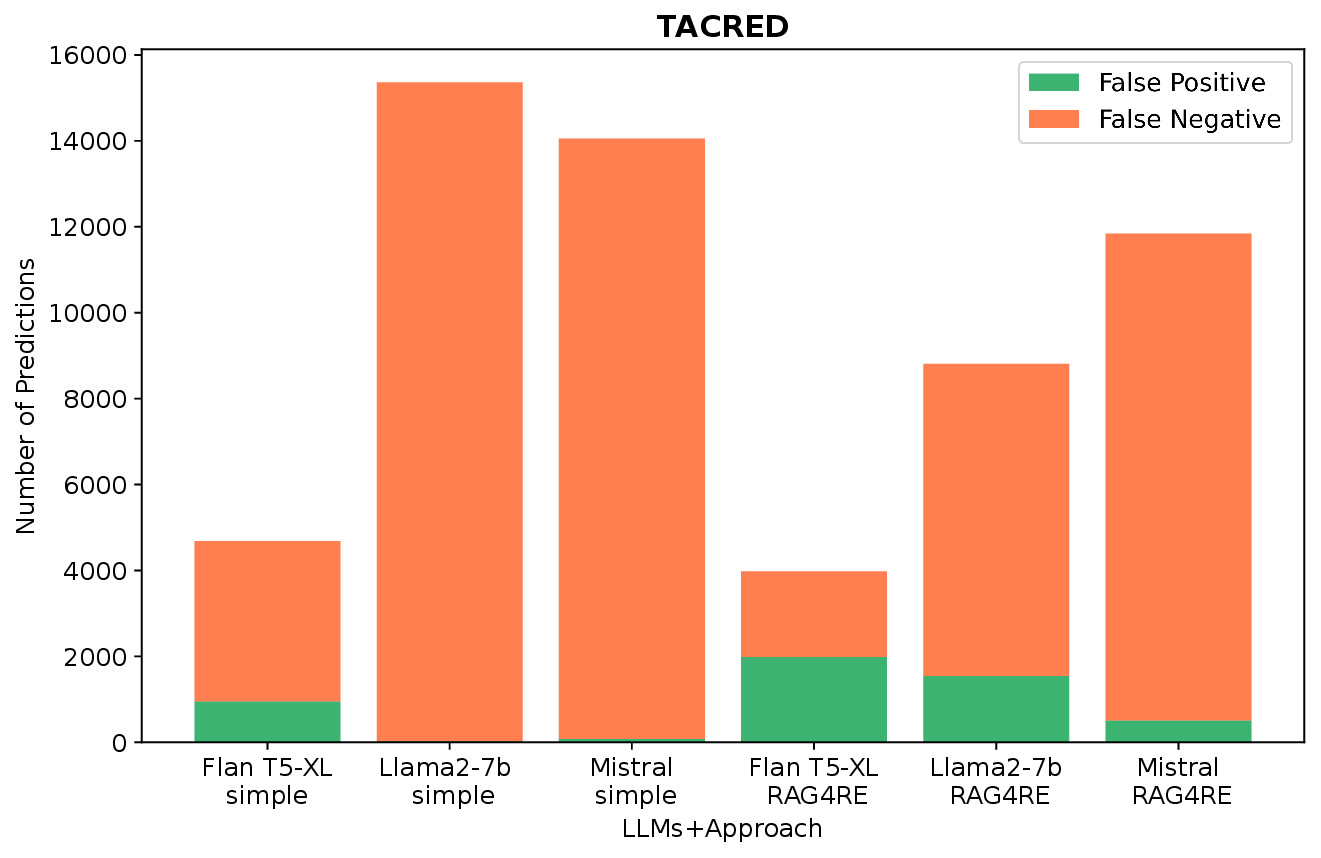} 
    \vspace{4ex}
  \end{minipage}
  \begin{minipage}[b]{0.62\linewidth}

    \includegraphics[bb=0 0 200 500, height=0.5\linewidth]{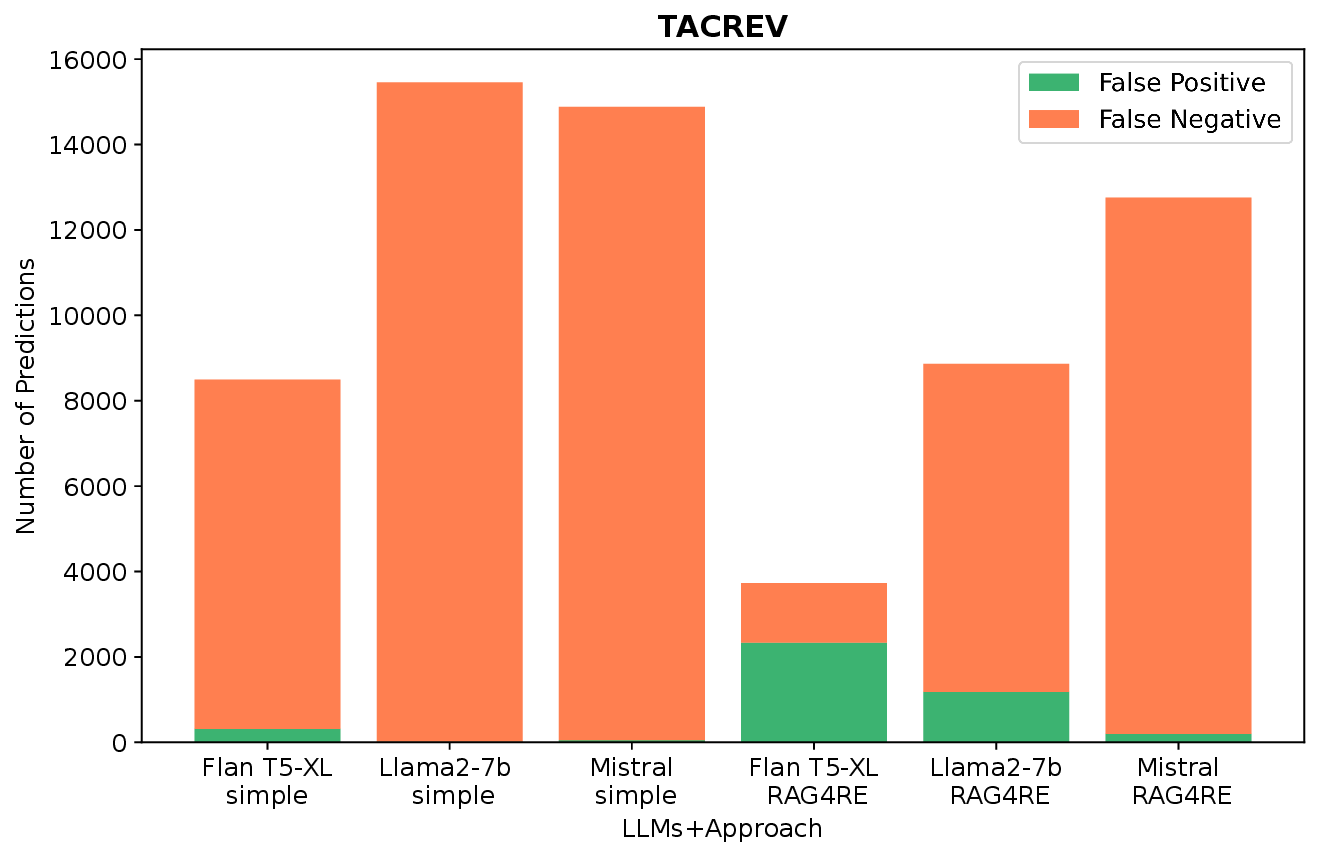} 
    \vspace{4ex}
  \end{minipage} 
  \begin{minipage}[b]{0.62\linewidth}
  
    \includegraphics[bb=0 0 200 500, height=0.5\linewidth]{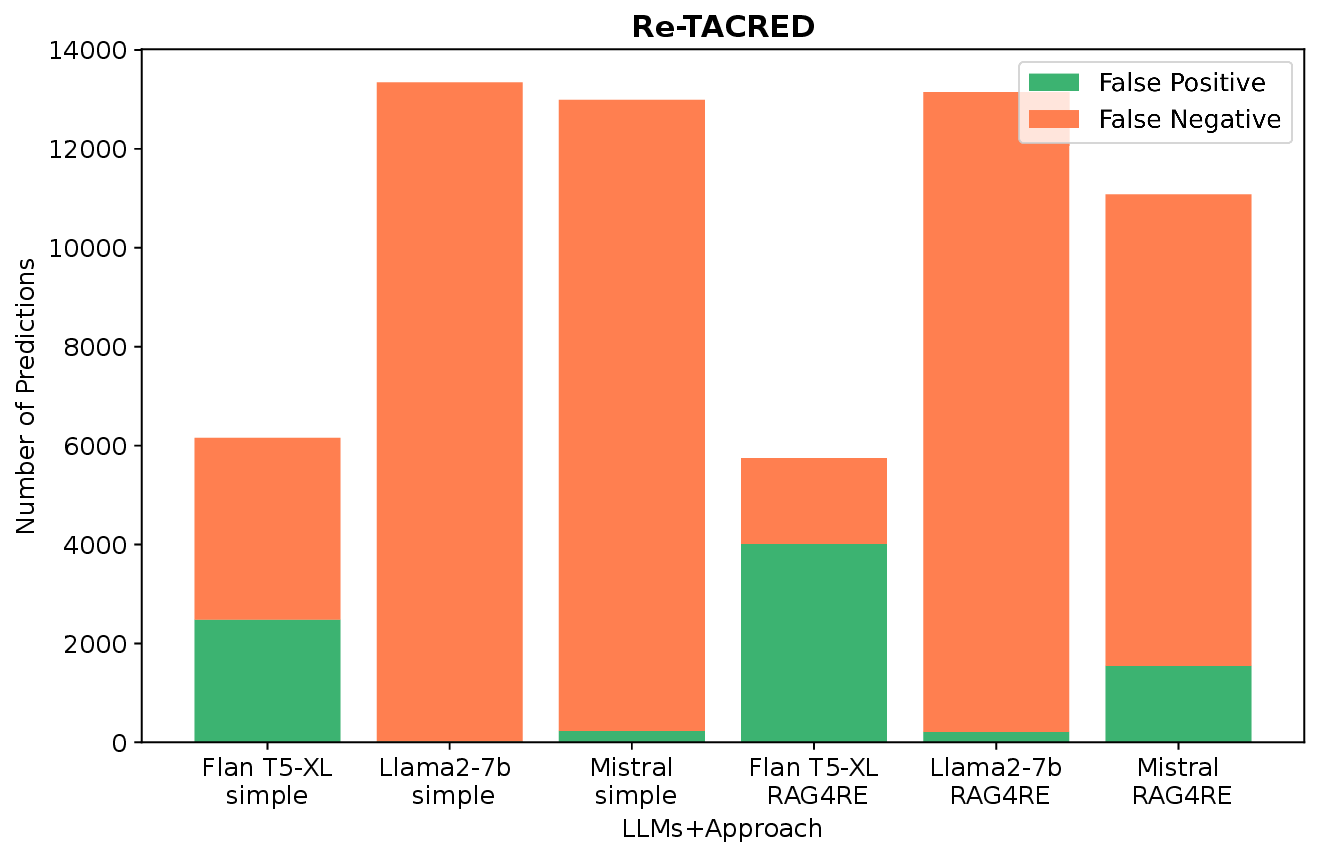} 
    \vspace{4ex}
  \end{minipage}
  \begin{minipage}[b]{0.62\linewidth}
    
    \includegraphics[bb=0 0 200 500, height=0.5\linewidth]{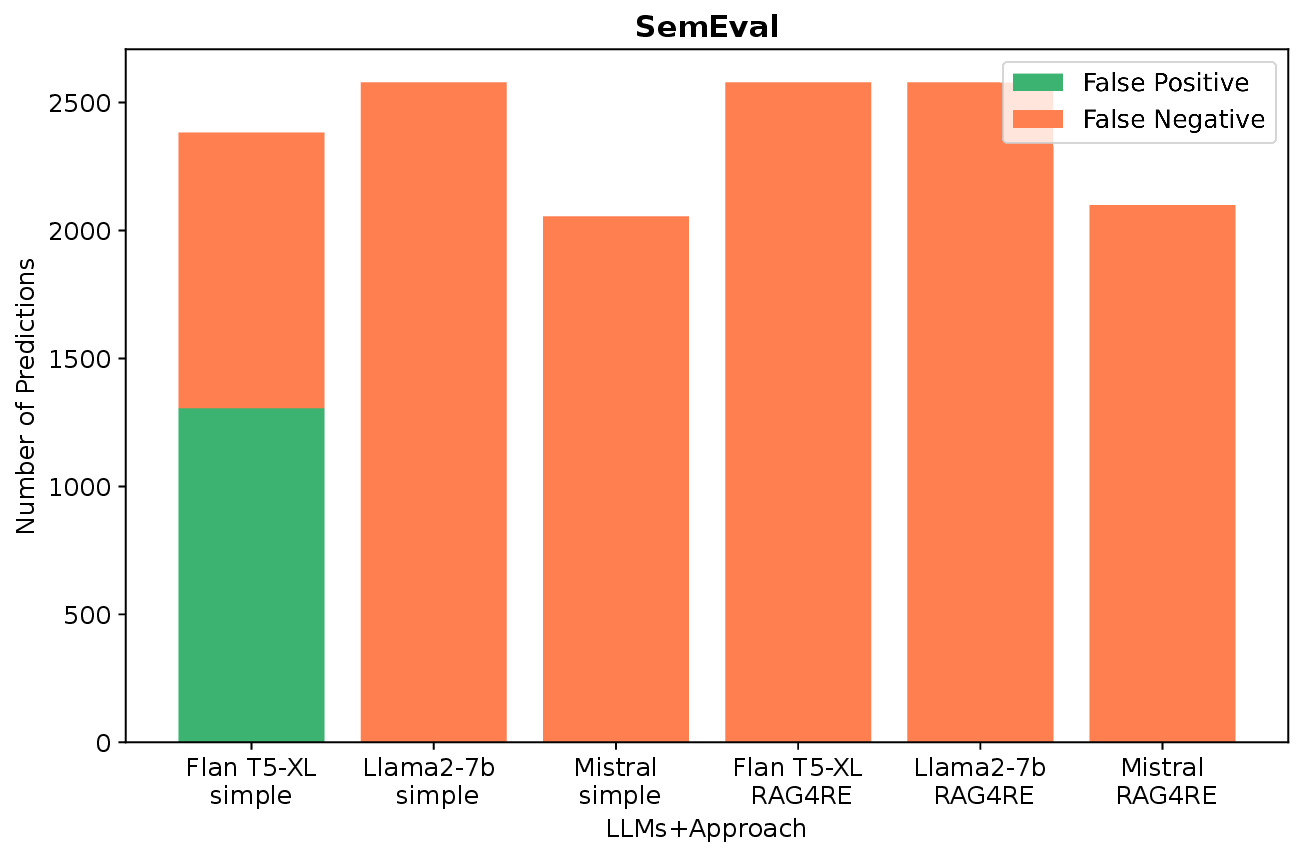} 
    \vspace{4ex}
  \end{minipage}

\end{figure}

Decoder only LLMs, such as Mistral and Llama2, are prone to producing hallucinatory results when a simple prompt is sent to those models~\cite{pan2024unifying}. In the responses of the experiments conducted with Mistral-7B-Instruct-v0.2 and Llama2-7b, we observed such relation types which have not been defined in the relation types of the prompt template (See the example prompt in~\Cref{fig:prompt_tempt}). We analyze False Negative and False Positive relation predictions of three language model types in~\Cref{tab:error_analysis} and~\Cref{fig:err}.

Regarding false predictions, our RAG4RE has decreased the number of false prediction in all experiments except for the SemEval dataset when the T5 model was integrated into the system. Likewise, Llama2-7b and Mistral-7B-Instruct-v0.2 illustrates a similar decreasing trend in the number of its false predictions.

In terms of False Negative (FN) predictions, our RAG4RE has decreased its FN predictions on the T5 model. Similarly, our RAG4RE has reduced the number of FN predictions on Mistral-7B-Instruct-v0.2 and  Llama2-7b.

Overall, when comparing the number of false responses, both Llama2-7b and Mistral-7B-Instruct-v0.2 produce higher numbers compared to Flan T5 XL. It is clear that RAG4RE has mitigate hallucination problem of the simple query (Vanilla LLM prompting) by reducing the number of false prediction on three language model types.

\begin{table}[H]
  \centering
  \caption{The table gives details about False Negative and Positive LLMs' responses.}
  \label{tab:error_analysis}
  \begin{tabular}{ l |l| r r |r r |r r  |r r }
    \toprule
     \multicolumn{2}{c}{\textbf{}} & \multicolumn{2}{c}{\textbf{TACRED}} &\multicolumn{2}{c}{\textbf{TACREV}}&\multicolumn{2}{c}{\textbf{Re-TACRED}} &\multicolumn{2}{c}{\textbf{SemEval}} \\
    \cmidrule{1-10}
    \multicolumn{1}{c}{\textbf{LLM}} & \multicolumn{1}{c}{\textbf{Approach}} & \multicolumn{1}{c}{FP} & \multicolumn{1}{c}{FN}&\multicolumn{1}{c}{FP} & \multicolumn{1}{c}{FN}&\multicolumn{1}{c}{FP} & \multicolumn{1}{c}{FN}&\multicolumn{1}{c}{FP} & \multicolumn{1}{c}{FN}\\
    \midrule
    \textit{Flan T5-XL} &simple query  &$956$&$3726$&$316$& $8180$& $2481$&$3676$&$1306$&$1076$\\
      & RAG4RE & $1990$ & $1987$& $2337$ & $1390$&$4013$&$1735$&$1564$ & $829$\\
      \hline
      \textit{Llama2-7b}&simple query&26&15336& 18&15438&18&13327&$0$ &$2579$ \\
      & RAG4RE & 1549&7258& 1184&7676&212&12937 & $0$ & $2603$\\
      \hline
    \textit{Mistral-7B-Instruct-v0.2} & simple query& 82 &13970 &52&14830&233&12759& $0$ & $2055$\\
         & RAG4RE&509&11333 &200&12557& 1546&9535& $0$ & $2099$ \\
    \bottomrule
  \end{tabular}
\end{table}

%% file: sections/discussion.tex
\subsection{Discussion}
In our experiments, we compared two methods: (i) RAG-based Relation Extraction (RAG4RE) and (ii) a simple query (Vanilla LLM prompting) which lacks the inclusion of the relevant example sentence described in~\Cref{fig:prompt_tempt}. Our findings indicate a notable enhancement in F1 scores when employing the RAG4RE approach over the simple query method on the TACRED dataset and its variations, as outlined in~\Cref{tab:sota_results}. This improvement stems from the integration of a relevant sentence into the prompt template, as illustrated in~\Cref{fig:prompt_tempt}. This inclusion of relevant example sentence facilitates the predictions made by the LLM in the Generation module of our proposed architecture, as depicted in~\Cref{fig:re-diagram}.

The incorporation of this relevant sentence serves to mitigate hallucinations in the LLM's responses, subsequently reducing the occurrence of false predictions, as demonstrated in~\Cref{tab:error_analysis} and ~\Cref{fig:err}. Our assessment of the RAG4RE approach's effectiveness is based on the integration of Flan T5-XL into the LLM within~\Cref{fig:re-diagram}, given that our approach, combined with Flan T5-XL, yields the highest F1 scores across benchmarks except for the SemEval (See~\Cref{tab:sota_results}).

We present an analysis of the performance of our RAG4RE approach, comparing its F1 score with both LLM-based methods and State-of-the-Art (SoTA) Relation Extraction (RE) techniques reported in current literature. In terms of LLM-based RE approaches, our RAG4RE consistently outperforms other methods utilizing LLMs, as illustrated in~\Cref{tab:llm}, across all benchmark datasets except for SemEval. The reason for the superior performance of our RAG4RE, as depicted in \Cref{tab:llm}, is largely attributed to the absence of relevant sentence addition in the prompt templates of both LLMQA4RE~\cite{Zhang2023LLM-QA4RE} and RationaleCL~\cite{xiong-etal-2023-rationale}. Notably, both competing methods did not incorporate relevant sentences into their prompt templates.

Furthermore, we conducted a comparison with works achieving the best results on these benchmarks in the literature. Remarkably, our RAG4RE surpasses the State-of-the-Art results on both TACRED and TACREV datasets, achieving F1 scores of 86.8\% and 88.3\% respectively, as illustrated in~\Cref{tab:sota}.

\begin{table}[H]
  \centering
  \caption{Comparing our best-performing results with State-of-the-Art (SoTA) Systems' results.}
  \label{tab:sota}
  \begin{tabular}{ l| r |r |r  |r }
    \toprule
    \textbf{Approach}& \textbf{TACRED}&\textbf{TACREV}&\textbf{Re-TACRED} &\textbf{SemEval} \\
 
    \midrule
     DeepStruct~\cite{wang-etal-2022-deepstruct}&$76.8\%$&-&-&-\\
     \hline
     EXOBRAIN~\cite{zhou-chen-2022-improved} &$75.0\%$&-&\textbf{91.4\%}&-\\
     \hline
     KLG~\cite{li2022reviewing}&- &$84.1\%$&-&$90.5\%$\\
     \hline
     SP~\cite{cohen2020relation} &$74.8\%$ &-& &\textbf{91.9\%}\\
     \hline
      GAP~\cite{CHEN2024123478} &$72.7\%$ &82.7\%& \textbf{91.4\%} &$90.3\%$\\
     \hline \hline \textbf{Ours (RAG4RE)}&\textbf{86.6\%}&\textbf{88.3\%}&$73.3\%$&$23.41\%$\\
    \bottomrule
  \end{tabular}
\end{table}
\begin{table}[H]
  \centering
  
  \caption{Comparing RAG4RE to Relation Extraction approaches using Language Models (LLMs)}
  \label{tab:llm}
  \begin{tabular}{ l| r |r |r  |r }
    \toprule
    \textbf{Approach}& \textbf{TACRED}&\textbf{TACREV}&\textbf{Re-TACRED} &\textbf{SemEval} \\
 
    \midrule
     LLMQA4RE~\cite{Zhang2023LLM-QA4RE}&$52.2\%$&$53.4\%$&$66.5\%$&$43.5\%$\\
     \hline
     RationaleCL~\cite{xiong-etal-2023-rationale}&$80.8\%$&-&-&-\\
     \hline \hline
     \textbf{Ours (RAG4RE)}&\textbf{86.6\%}&\textbf{88.3\%}&\textbf{73.3\%}&$23.41\%$\\
    \bottomrule
  \end{tabular}
\end{table}

Overall, our RAG4RE demonstrates impressive performance on the TACRED and TACREV datasets. However, its performance on SemEval did not reach similar improvement, primarily due to challenges posed by predefined relation types (target relation labels) in this particular benchmark dataset. For instance, directly extracting the ``Cause-Effect (e2,e1)" relation type from the provided sentence tokens between entity 1 (e1) and entity 2 (e2) remains a challenge for zero-shot LLM prompting, as this type of relation often requires logical inference to be accurately identified.

%% file: sections/conclusion.tex
\section{Conclusion and Future Work}

In this work, we introduce a novel approach to Relation Extraction (RE) called Retrieval-Augmented Generation-based Relation Extraction (RAG4RE). Our aim is to identify the relation types between head and tail entities in a sentence, utilizing an RAG-based LLM prompting approach.

We also claim that RAG4RE has outperformed the performance of the simple query (Vanilla LLM prompting). To prove our claim, we conducted experiments using four different RE benchmark datasets: TACRED, TACREV, Re-TACRED, and SemEval, in conjunction with three distinct LLMs: Mistral-7B-Instruct-v0.2, Flan T5, and Llama2-7b. Our RAG4RE yielded remarkable results compared to those of the simple query (Vanilla LLM prompting). Our RAG4RE exhibited notably results on the benchmarks compared to previous works. Unfortunately, our proposed methods did not perform on the SemEval dataset. This is due to the lack of logical inference in LLMs, as predefined or target relation
types cannot be directly extracted from sentence tokens on the SemEval dataset.

 In our future work, we plan to extend our approach to real-world dynamic learning scenarios and evaluate it using real-world datasets. Additionally, we intend to integrate fine-tuned LLMs on training datasets into our RAG4RE system to address the performance issues encountered when datasets require logical inference to identify relation types between entities in a sentence, and target relation types cannot be extracted from the sentence tokens as in SemEval.